\documentclass[sigplan,screen,10pt]{acmart}
\AtBeginDocument{%
  \providecommand\BibTeX{{%
    Bib\TeX}}}

\usepackage{algorithm}
\usepackage{algorithmicx}
\usepackage[noend]{algpseudocode}
\usepackage{multirow}
\usepackage{booktabs}

\usepackage{graphicx}
\usepackage{textcomp}
\usepackage{xcolor}
\usepackage{hyperref}
\usepackage{tabularray}

\hypersetup{hypertex=true}
\def\BibTeX{{\rm B\kern-.05em{\sc i\kern-.025em b}\kern-.08em
    T\kern-.1667em\lower.7ex\hbox{E}\kern-.125emX}}

\newcommand{\method}{AdapMoE}
\usepackage{etoolbox}
\makeatletter
\patchcmd{\authornote}{\g@addto@macro\addresses{\@authornotemark}}{}{}{}
\makeatother

\copyrightyear{2024} 
\acmYear{2024} 
\setcopyright{acmlicensed}\acmConference[ICCAD '24]{IEEE/ACM International
 Conference on Computer-Aided Design}{October 27--31, 2024}{New York, NY,
 USA}
 \acmBooktitle{IEEE/ACM International Conference on Computer-Aided Design
 (ICCAD '24), October 27--31, 2024, New York, NY, USA}
 \acmDOI{10.1145/3676536.3676741}
 \acmISBN{979-8-4007-1077-3/24/10}

\begin{document}

\title{AdapMoE: Adaptive Sensitivity-based Expert Gating and Management for Efficient MoE Inference}

\author{Shuzhang Zhong$^{1,2}$, Ling Liang$^{2}$, Yuan Wang$^{2,4}$, Runsheng Wang$^{2,3,4}$, Ru Huang$^{2,3,4}$, Meng Li$^{1,2,4*}$}
\authornote{Corresponding Author, meng.li@pku.edu.cn}
\affiliation{%
  \institution{$^1$Institute for Artificial Intelligence \& $^2$School of Integrated Circuits, Peking University, Beijing, China}\country{}
}
\affiliation{%
  \institution{$^3$Institute of Electronic Design Automation, Peking University, Wuxi, China}\country{}
}
\affiliation{%
  \institution{$^4$Beijing Advanced Innovation Center for Integrated Circuits, Beijing, China}\country{}
}


\begin{abstract}
    Mixture-of-Experts (MoE) models are designed to enhance the efficiency of large language models (LLMs) without proportionally increasing the computational demands. However, their deployment on edge devices still faces significant challenges due to high on-demand loading overheads from managing sparsely activated experts. 
    This paper introduces \method, an algorithm-system co-design framework for efficient MoE inference.
    \method~features adaptive expert gating and management to reduce the on-demand loading overheads. 
    We observe the heterogeneity of experts loading across layers and tokens, based on which we propose a sensitivity-based strategy to adjust the number of activated experts dynamically. Meanwhile, we also integrate advanced prefetching and cache management techniques to further reduce the loading latency. Through comprehensive evaluations on various platforms, we demonstrate \method~consistently outperforms existing techniques, reducing the average number of activated experts by 25\% and achieving a 1.35$\times$ speedup without accuracy degradation. Code is available at: \href{https://github.com/PKU-SEC-Lab/AdapMoE}{https://github.com/PKU-SEC-Lab/AdapMoE}.
\end{abstract}

\maketitle
\pagestyle{fancy}
\fancyhead{} 

\section{Introduction}


Mixture-of-Experts (MoE) \cite{shazeer2017outrageously} models are designed to relieve the high compute requirements of large language models (LLMs) by introducing sparsely activated pathways.
The feedforward block in MoE uses a gating function to select from different parameter groups, allowing the model to scale in size without a corresponding increase in computation.
For instance, in the Mixtral-8x7b model, each token can access 47 billion parameters, yet only 13 billion active parameters are utilized during each inference iteration \cite{jiang2024mixtral}.

Unfortunately, the substantial memory requirements of MoE makes it difficult to be deployed, especially on edge devices with limited memory. A common solution is offloading, where expert weights are stored on CPU memory, SSDs, or Flash storage and loaded into the GPU only when needed \cite{aminabadi2022deepspeed,ren2021zero,shazeer2017outrageously}. However, the approach of dynamically loading activated experts on-demand leads to the serialization of the expert selection phase with the expert execution phase, thereby introducing significant performance overhead.

\begin{figure}[!tb]
    \centering
    \includegraphics[width=0.9\linewidth]{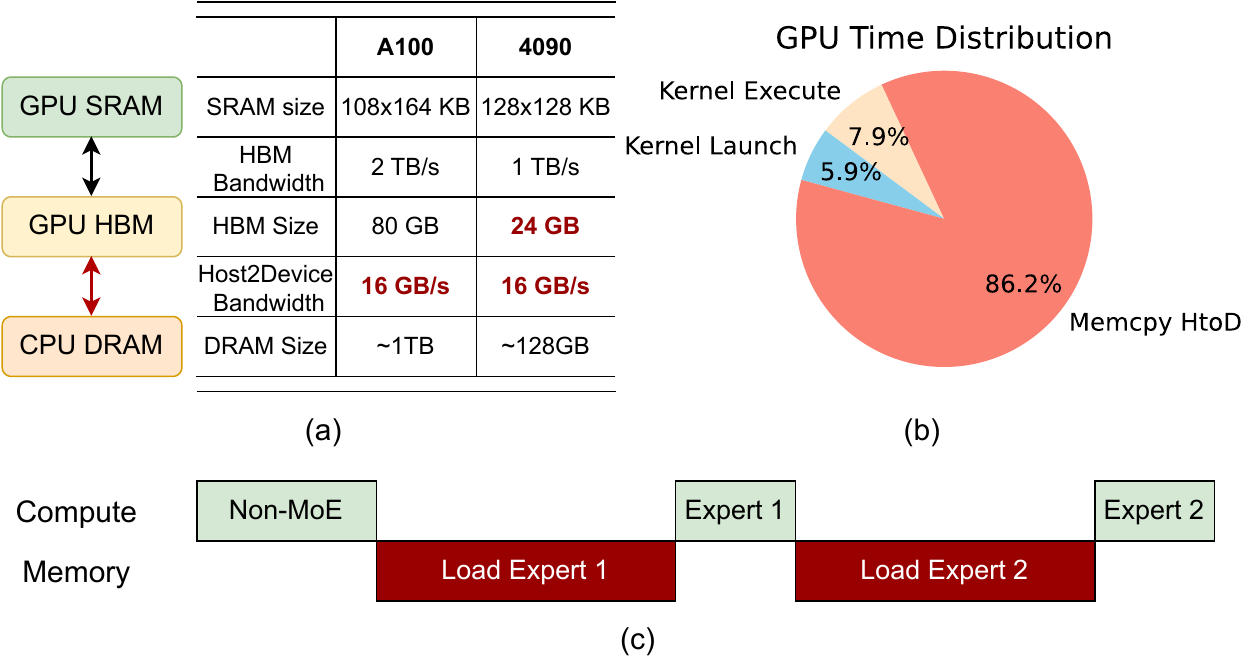}
    \caption{GPU resource utilization and task execution timeline: (a) hardware specs comparison between NVIDIA A100 and 4090 GPUs; (b) GPU time distribution; (c) task timeline with offloading.}
    \label{fig:intro}
    \vspace{-3mm}
\end{figure}

The main bottleneck in MoE inference is attributed to the \textbf{on-demand loading} of experts. As illustrated in Figure \ref{fig:intro}, this loading phase constitutes the majority of the latency during MoE inference, primarily due to the substantial communication scale and limited bandwidth available.

Faced with these challenges, existing works have explored several efficient MoE inference techniques.
Adap-gating \cite{li2023adaptive} allows tokens to be processed by a variable number of experts based on expert probability distribution. Though efficient, the heuristic gating criteria will lead to low accuracy compared with original fixed top-k routing. 
Mixtral-offloading \cite{eliseev2023fast} observes token-wise expert selection similarities and adopts least recently used(LRU) algorithm to manage the expert cache. Its fixed cache allocation strategy across each layer limits its adaptability and effectiveness.
MoE-Infinity \cite{xue2024moe} observes correlations in expert selection across layers, using previous layers' activations to predict subsequent expert activations. However, its straightforward predictive approach results in lower prefetching accuracy.
Pre-gated MoE \cite{hwang2023pre} modifies the structure of MoE, in which the experts are selected based on the previous layers' activation. Despite its ability to accurately predict expert usage and execute prefetching, the structural modifications will complicate direct deployment.

Despite these advances in MoE inference techniques, all existing solutions fail to comprehensively address all the critical factors affecting system performance simultaneously. Specifically, there has been a lack of a formal optimization approach that integrates expert gating, prefetching, and caching within a unified framework. Adap-gating and Pre-gated MoE will even cause accuracy degradation if directly deployed without finetuning. 

In this work, we propose AdapMoE, an algorithm-system co-design framework that enables adaptive expert gating and management for MoE inference without accuracy degradation. We identify three critical factors that significantly influence the on-demand loading overhead: the number of activated experts, the accuracy of prefetching, and the cache hit probability. Targeting these factors, our contributions in designing AdapMoE include:
\begin{itemize}
    \item \textbf{Adaptive and sensitivity-based expert gating.} 
    We propose an adaptive and sensitivity-based gating mechanism for the MoE inference process, which allows tokens to be processed by a dynamically flexible number of experts without accuracy degradation.
    \item \textbf{Adaptive expert prefetching.} We observe the high similarities between activations in different layers and leverages the gating decisions of subsequent layers as predictive signals to prefetch experts for these layers in advance.
    \item \textbf{Adaptive cache allocation.} We proposed a dynamic programming based formulation for the cache size allocation to minimize the on-demand loading overhead.
    \item \textbf{System implementation.} We implement expert management of AdapMoE with fine-grained CUDA stream controller to maximize the overlap between communications and computations. 
\end{itemize}

We evaluate the performance of AdapMoE system using state-of-the-art Mixtral models on various platforms and datasets. Our adaptive gating algorithm can maintain high accuracy levels while significantly reducing the experts activation by 25\% compared with traditional top-2 gating. AdapMoE consistently outperforms all the existing expert management methods by 1.35x.

\section{Background}

\subsection{Mixture-of-Experts}
Mixture-of-Experts(MoE) \cite{masoudnia2014mixture}  models present a solution to manage the extensive computational demands of LLMs through sparsely activated pathways, including Mixtral \cite{jiang2024mixtral}, NLLB-MoE \cite{costa2022no}, and SwitchTransformer \cite{fedus2022switch}. 
A gating function $G(x)$ determines which of several distinct parameter groups, or "experts," are activated for a given input:
\begin{align}
    G(x) = \text{Softmax}(\text{TopK}(x\cdot W_g))
\end{align}

The MoE module outputs are determined by a weighted sum of individual expert outputs. Given N experts $E_0, E_1$,..., $E_{N-1}$, the output can be expressed as:
\begin{align}
    \sum_{i=0}^{N-1}G(x)_i\cdot E_i(x),
\end{align}
The total number of experts N and the number of activated experts K vary among different MoE implementations. For instance, the Mixtral model employs 8 experts, with only 2 being active at a time.

\subsection{Efficient MoE Inference}
The substantial memory requirements of MoE makes it difficult to be deployed on edge devices with limited memory. To handle this problem, a typical solution is to offload model parameters into CPU memory or SSDs, such as DeepSpeed \cite{aminabadi2022deepspeed} and FlexGen \cite{sheng2023flexgen}. However, they are designed for dense models and will load/prefetch all parameters, leading to unnecessary communication overhead.

In response to the specific challenges posed by sparse expert activation in MoE models, various specialized MoE inference techniques have been developed, including Adap-gating\cite{li2023adaptive}, EdgeMoe \cite{yi2023edgemoe}, MoE-Infinity \cite{xue2024moe}, Mixtral-offloading \cite{eliseev2023fast} and Pre-gated MoE \cite{hwang2023pre}. 
These frameworks propose several gating, prefetching or caching algorithm. Despite these advancements, many of these solutions still struggle with critical issues. For instance, frameworks like Adap-gating and Pre-gated MoE can experience significant accuracy degradation when directly deployed. 
These techniques fail to comprehensively address all key factors affecting accuracy and performance, which leads to suboptimal performance.

\section{Motivations and observations}
\label{sec:motivation}
The primary bottleneck in MoE edge inference is attributed to the on-demand loading of experts, a process that cannot be fully overlapped with computation and thus occupies a significant portion of the total inference time. To address this issue, we begin by analyzing the main factors that influence the frequency of on-demand loading:

 \textbf{Number of activated experts:} Reducing the number of activated experts can directly decrease the frequency of on-demand loading.
    
\textbf{Accuracy of prefetching:} Accurate prefetching allows the overlap of expert loading with computation in subsequent layers, effectively eliminating performance impacts. 

 \textbf{Cache hit probability:} If the required experts are already cached in memory, computations can be performed directly without on-demand loading. 

Targeting these factors, we make the following observations that motivate us to propose AdapMoE:

\begin{figure}[!tb]
    \centering
    \includegraphics[width=\linewidth]{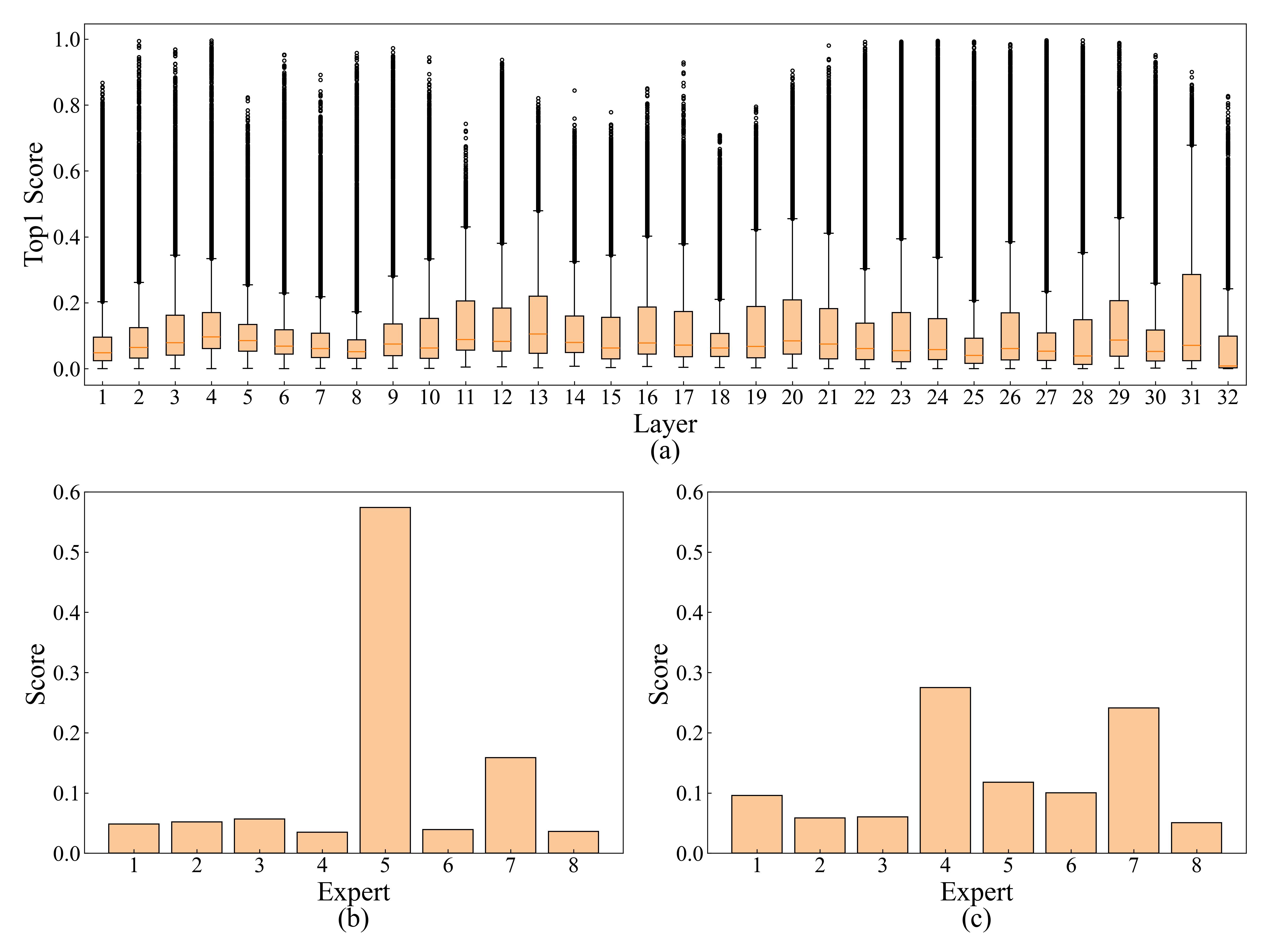}
    \caption{Expert weight score distribution: (a) scores of top-1 expert per layer; (b) (c) weight score distribution examples. }
    \label{fig:distribution}
    \vspace{-3mm}
\end{figure}

\textbf{Observation 1: Dynamic Expert Demand.} Intuitively, the requirement for experts varies across different tokens and layers. As illustrated in Figure \ref{fig:distribution}, some tokens exhibit a significantly biased weight distribution, necessitating fewer experts. Additionally, the relevance of different layers to the final output varies; layers of less importance can tolerate a reduction in the number of activated experts without substantial impact on the overall model performance. By activating fewer experts, the demand for on-demand loading overheads can be directly reduced.

\begin{figure}[!tb]
    \centering
    \includegraphics[width=0.9\linewidth]{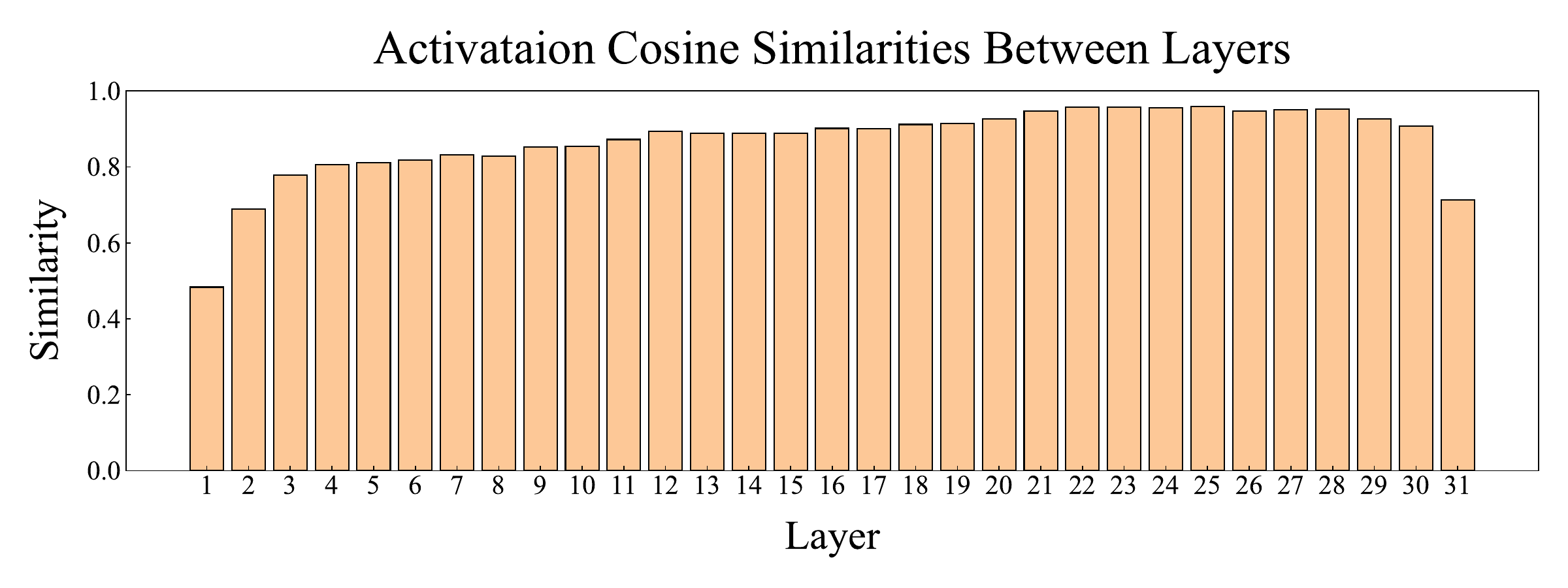}
    \caption{Cosine similarities between the input of each layer's moe block and the input of its next layer's moe block.}
    \label{fig:similarity}
\end{figure}

\textbf{Observation 2: Similar intermediate activation.} As shown in figure~\ref{fig:similarity}, the activations between successive layers exhibit a very high degree of similarity. This characteristic is due to the nature of the residual connections, which will accumulate the output of each layer, thus maintaining a consistency in the feature space as the depth increases. The similarity can be utilized to predict the experts needed in upcoming layers more accurately.

\textbf{Observation 3: Unbalanced expert cache demands.} Different layers varies in the prefetching accuracy and required number of activated experts, leading to adaptive demands for cache size under different platforms. Effective cache management must adapt to these varied needs to optimize overall system performance, ensuring resources are used where they are most needed.

\section{AdapMoE Design}

\subsection{Overview}
\begin{figure}
    \centering
    \includegraphics[width=\linewidth]{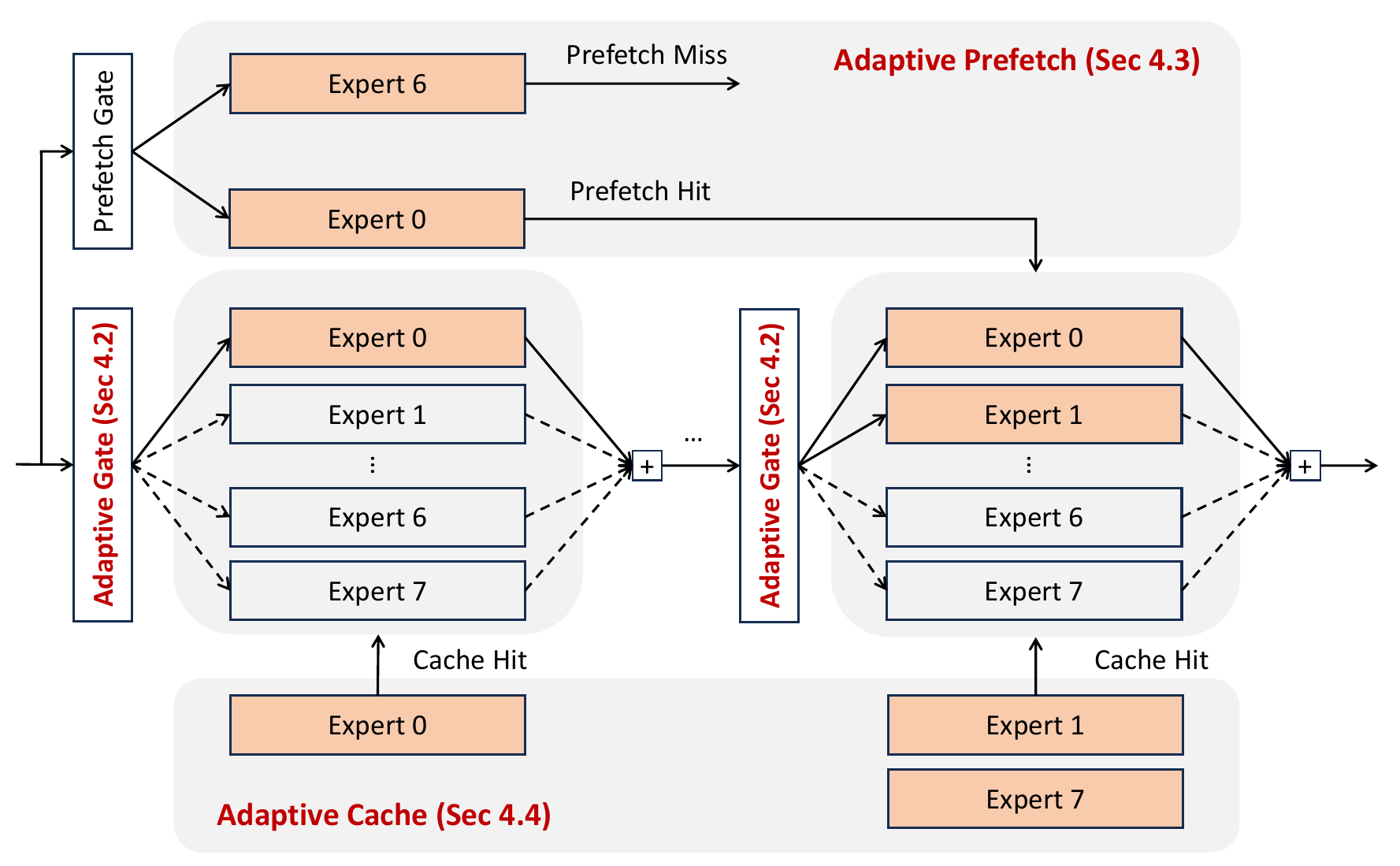}
    \caption{Overview of AdapMoE.}
    \label{fig:overview}
\end{figure}

This paper introduces AdapMoE, an algorithm-system co-design for scalable and efficient MoE inference on edge devices.
AdapMoE is designed to address the high latency overhead associated with the on-demand loading of experts during MoE inference.
AdapMoE proposes 
(i) a sensitivity-based adaptive gating mechanism that allows tokens to be processed by a dynamically flexible number of experts(Sec. \ref{gating}), 
(ii) an expert prediction and adaptive prefetching algorithm that achieves high prefetching accuracy without modification and finetuning of models(Sec. \ref{prefetching}), 
 and (iii) a DP-based cache allocation algorithm to meet the unbalanced expert cache demands(Sec. \ref{caching}).

Figure~\ref{fig:overview} presents an architectural overview of AdapMoE. 
The system initially collects sensitivity and prediction data from a sampled dataset in the offline phase. Subsequently, it applies adaptive gating, prefetching, and caching during online operation. Adaptive gating will influence the prefetching algorithm, and both gating and prefetching will then guide caching decisions, optimizing resource allocation to match real-time data variability effectively.

\subsection{Adaptive and Sensitivity-based Expert Gating} \label{gating}
As we plot in figure~\ref{fig:distribution}, the score distribution of experts differs among different tokens and different layers. The distribution clearly demonstrate an unbalanced pattern, where the required number of experts are different during inference. Therefore, the main task for adaptive expert gating is to find an optimal way to allocate distinct number of experts for each token in each layer. 

Thus, it is crucial to quantify the sensitivity of the gating mechanism to fluctuations in the number of experts per layer. We define a perturbation metric that measures the deviation in performance when varying the number of active experts. This metric allows us to establish a threshold $T$. If the perturbation is less than the threshold $T$, the gating system can safely activate a smaller subset of experts.



Let $x$ be the input of MoE block, $f_1$ be the expert with highest score, $f_2$ be the expert with the second highest score, $\alpha$ be the score of Top-1 expert after normalization in layer $i$. The original output $O_0$ of the sparse moe block in layer $i$ can be represented as:
\begin{align}
    O_0 = \alpha f_1(x) + (1-\alpha)f_2(x) 
\end{align}

If we conduct adaptive gating and only activate the first expert, the output O can be represented as:
\begin{align}
    O = f_1(x)
\end{align}

To determine the influence of adaptive gating in each layer, we can perform Taylor series expansion to analyze how the model output changes in response to perturbations by the adaptive gating.
\begin{align}
    L(O)& \simeq L(O_0) + g^T(O-O_0) + \frac{1}{2}(O-O_0)^T H (O-O_0)  \\
        & \simeq L(O_0) + \frac{1}{2} (1-\alpha)^2 (f_1(x)-f_2(x))^T H (f_1(x)-f_2(x)) \nonumber
\end{align}
where g is the gradient of $O_0$ and $H = E[\frac{\partial ^2}{\partial O_0^2}L(O_0)]$ is the second derivative (i.e, Hessian) of the loss at $O_0$. Assuming the model has converged to a local minimum, we can approximate the gradient $g$ as being zero \cite{kim2023squeezellm}.

Hessian matrices can be computed using a second back-propagation \cite{yao2020pyhessian}; however, this approach is highly memory-intensive, making it impractical, especially for MoE models.  To address these challenges, we adopt Fisher information matrix to approximate the hessian matrix following \cite{kim2023squeezellm}.
\begin{align}
    H \simeq F = \frac{1}{|D|} \sum_{d\in D} g_d g_d^T,
\end{align}
where $D$ is the sample dataset.

The term $f_1(x)-f_2(x)$ represents the difference between two experts. However, it is input-specific and hard to be measured. What's more, difference of  experts in a well-trained MoE model should be similar. Thus we approximate this term as:
\begin{align}
    &(f_1(x)-f_2(x))^T H (f_1(x)-f_2(x) \nonumber) \\
    \simeq &(f_1(x)-f_2(x))^T diag(F) (f_1(x)-f_2(x)) \propto \sum diag(F)
\end{align}
Finally, given a threshold $T$, only one expert will be activated if the Hessian and weight score in this layer satisfy:
\begin{align}
    (1-\alpha)^2 \times \sum diag(F) \leq T,
\end{align}
where $diag(F)$ denotes the diagonal elements of $F$. The value of $T$ is uniformly applied across all layers in the MoE model. The specific value of $T$ is determined by identifying the optimal balance on a validation set, where it minimizes the average number of expert activations without compromising the model’s accuracy. Compared with score-based adaptive gating \cite{li2023adaptive} that only takes $\alpha$ into account, our sensitivity-based gating considers the layer importance and allows for higher single expert activation ratios without compromising accuracy.

\subsection{Adaptive Expert Prefetching} \label{prefetching}

Expert prefetching can significantly mitigate on-demand loading overheads by overlapping the communication latency with computation of MoE and attention blocks.
Meanwhile, the prefetching efficiency relies highly on the prediction accuracy of next layer's expert selection.

\begin{figure}[!tb]
    \centering
    \includegraphics[width=\linewidth]{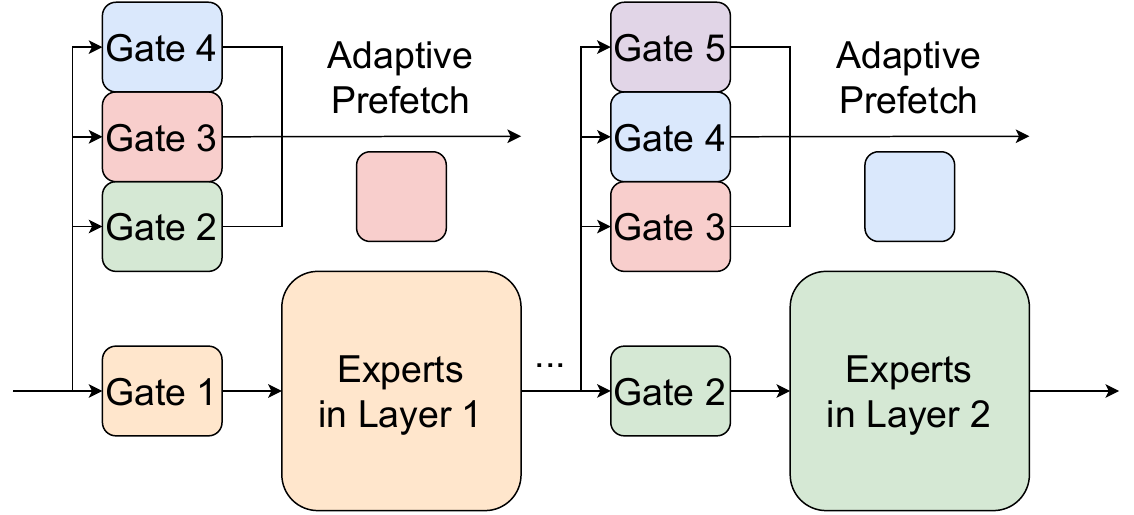}
    \caption{Adaptive prefetching workflow. Gate i represents the gating function in layer i. Adaptive gating enables prefetch multiple next layers' experts.}
    \label{fig:prefetch}
\end{figure}

As depicted in figure~\ref{fig:prefetch}, AdapMoE utilizes the gate functions from the subsequent layer directly for predictive prefetching. The key to this adaptive prefetching is the high similarity in activations across layers, as established in Observation 2 from Section \ref{sec:motivation}.

Furthermore, the activations are indicative of the expert selections not only for the immediate next layer but potentially for the next two/three layers. AdapMoE predicts potentially selected experts for the subsequent layers, incorporating adaptive gating into predictions. If the experts needed by the next layer are already cached in the GPU memory, AdapMoE preemptively fetches experts required for subsequent layers, extending beyond the immediate next. 

However, one important question still reamains: How do we prefetch the experts for the first layer?
Given that the first layer lacks preceding layers from which to infer gating information, previous works, such as Pre-gated MoE, are constrained to on-demand loading these experts.
Here we introduce an auxiliary predictive layer specifically designed to estimate the gating for the first layer's expert selection with the last layer activation of the previous token. This predictive layer is trained to minimize the divergence between its predictions and the actual gating decisions of the first layer, thereby allowing for a more efficient prefetching strategy that can potentially reduce latency significantly.

Let $D_{KL}$ denote the Kullback-Leibler(KL) divergence, $A_{last}$ represents the activations from the last layer of the previous token, $A_{first}$ denote the activations from the first layer of the current token, and $G_{first}$ be the gate function for the first layer, $G_{pre}$ be the predictive gate. The loss $L$ is computed as follows:
\begin{align}
    L=D_{KL}(softmax(G_{first}(A_{first}))_{[:,0:-1,:]} \nonumber \\
    ||softmax(G_{pre}(A_{last}))_{[:,1:,:]})
\end{align}

The training of this predictive gate is optional, providing flexibility based on resource availability computational constraints. Also, the training overhead for this gate is very samll, for the parameter count is only hidden\_dim $\times$ expert\_num.
Experiments demonstrate that the predictive gate dedicated to the first layer's expert selection, along with the reuse of gate functions in subsequent layers, both achieve high accuracy in prefetch, which will further elaborate in section \ref{sec:exp}.

\subsection{Adaptive Expert Caching} \label{caching}
Expert caching is critical for MoE edge inference. By storing frequently selected experts in the GPU memory, expert caching can significantly reduce the on-demand loading overheads.   

However, due to adaptive expert gating and adaptive prefetching, the cache demands for experts across different layers can become unbalanced. On the one hand, layers that tend to select more experts require a larger cache footprint to store these experts in the GPU memory; On the other hand, layers with lower prefetching accuracy also require increased cache storage to accommodate potential misses, ensuring that subsequent computation will not be stalled by on-demand loading delays.

In this section, we first quantify the on-demand loading overheads for each layer by analyzing tracing data, which encompasses metrics related to adaptive gating and prefetching. Then we formulate the cache size allocation problem as a knapsack problem. Finally, we employ a dynamic programming algorithm to find the optimal cache allocation strategies on given platforms. 

\subsubsection{Variable Definition}
\begin{table}[!tb]
    \centering
    \caption{Notations used for the DP formulation.}
    \begin{tabular}{cc}
    \hline \hline
        Variable & Meaning \\ \hline
        $T$ & Total expert cache size \\ \hline
        $L$ & Number of model layers \\ \hline
        $N$ & Number of experts per layer \\ \hline
        $t_{i}$ & Expert cache size of layer i \\ \hline
        $\alpha_{i}$ & Single expert gating probability \\ \hline
        $\beta_{i}$ & Prefetching accuracy of layer i \\  \hline
        \multirow{2}[0]{*}{$f_{i,t}$} & On-demand loading quantities \\
          & of layer i under cache size t \\
    \hline \hline 
    \end{tabular}
    \label{tab:var}
\end{table}

The notations of the DP formulation are defined in table~\ref{tab:var}. Among these variables, $T$ is predetermined by the hardware configuration, $\alpha$ and $\beta$ are determined by offline profiling, $t$ is the optimization target. The function $f_i^t$, representing the on-demand loading frequency for each layer under different cache sizes, can be quantified through these variables. The details of how $f_i^t$ is calculated will be introduced in the following section.

\subsubsection{On-demand Loading Quantification}
In this section, we use the Mixtral MoE model to analyze the representation of on-demand loading costs, which will select 2 experts from 8 experts per layer.

First, we calculate the probability of hitting the cache $p_{hit}$ of a specified expert. Due to the expert similarity between successive tokens is not significant, we can simplify the calculation of $p_{hit}$ by directly relating it to the cache size allocated for each expert.
\begin{align}
    p_{hit} = \frac{t_{i}}{N}
\end{align}
Similarly, the cache hit probability of two specified experts can be represented as $\frac{t_{i}\times t_{i}-1}{N\times (N-1)}$, the cache miss probability of two specified experts can be represented as $max(\frac{(N-t_{i})\times (N-t_{i}-1)}{N\times (N-1)},0)$.

We then explore the on-demand loading costs based on the number of activated experts:

\textbf{When only one expert is required.} The on-demand loading overhead occurs if there is a cache miss combined with incorrect prefetching. This cost is represented as:
        \begin{align}
            f_{i,t}^1 = (1-\frac{t_i}{N}) \times (1-\beta_{i})
        \end{align}

\textbf{When two experts are required.} This scenario introduces more complexity and can be broken down into several cases:
        \begin{itemize}
            \item Both experts miss the cache and prefetching is incorrect: In this case, both experts need to be loaded on-demand, which significantly increases the loading overhead. The loading cost can be represented as:
                \begin{align}
                    f_{i,t}^2 = 2 \times max(\frac{(N-t_{i})\times (N-t_{i}-1)}{N\times (N-1)},0) \times (1-\beta_i)
                \end{align}
            \item Both experts miss the cache but prefetching is correct for one expert: Here, only one expert needs to be loaded on-demand. The on-demand loading costs can be represented as:
                \begin{align}
                    f_{i,t}^3 = max(\frac{(N-t_{i})\times (N-t_{i}-1)}{N\times (N-1)},0) \times \beta_{i}
                \end{align}
            \item One expert hits the cache but prefetching is incorrect: The on-demand loading cost for this case can be calculated by considering the probability that one specific expert is cached and the other is not, yet the prefetching inaccurately predicts the need for the second expert. The formula for this cost is the following.
                \begin{align}
                    f_{i,t}^4 = \frac{2 \times (N-t_{i})\times t_i}{N \times (N-1)} \times (1-\beta_i)
                \end{align}
        \end{itemize}

Finally, the overall on-demand loading costs of layer i can be calculated as:
\begin{align}
    f_{i,t} = \alpha_i \times f_{i,t}^1  + (1-\alpha_i)\times(f_{i,t}^2+f_{i,t}^3+f_{i,t}^4)
\end{align}

\subsubsection{DP Formulation}
The goal of adaptive caching in MoE inference is to optimize the values of $t_i$ for each layer. This optimization aims to minimize the total on-demand loading overheads while staying within the limits of the total cache size $T$.

The problem can be mathematically formulated as an optimization scenario where:

\textbf{Objective Function.} Minimize the total expected on-demand loading cost, expressed as a sum of the individual costs $f_{i,t_i} $for each layer $i$.
        \begin{align}
            \min \quad \sum_i f_{i,t_i}
        \end{align}
 \textbf{Constraints.} Constraint \ref{eq:total} means the sum of the cache sizes allocated to each layer must not exceed the total available cache size; Constraint \ref{eq:single} ensures Each $t_i$ must be non-negative and cannot be greater than the total number of experts N at any layer.
        \begin{align}
           &\sum_{i=1}^L t_i \leq T \label{eq:total} \\
            &0\leq t_i \leq N \quad \forall i \label{eq:single}
        \end{align}
        
This can be viewed as a knapsack problem and solved by dynamic programming. Define $F[i][j]$ as the minimum on-demand loading cost achievable by considering the first $i$ layers and using up to j units of expert cache. The aim is to compute $F[L][T]$.

We start with $F[0][j]=0$ and then for each layer $i \in [1,L]$ and for each possible cache capacity $j \in [0,T]$, derive the optimal substructure of F as
\begin{align}
    F[i][j] = min_{k\leq j}(F[i-1][j-k] + f_{i,k})
\end{align}

To finalize the cache allocations, start from the last layer and total cache size, and trace back through each layer to determine the specific cache allocation that led to the minimum cost.

\section{System Implementation}

The implementation of the AdapMoE system is architected to take full advantage of GPU resources, specifically utilizing CUDA streams for efficient overlap of memory operations and computation.

The workflow, as outlined in Algorithm~\ref{algorithm}, begins by processing the experts that are already loaded into the GPU. It then efficiently handles experts that need to be loaded on-demand. By managing these tasks in parallel, the system ensures that computing resources are utilized effectively without waiting for data transfers to complete.

\begin{figure}[!tb]
    \centering
    \includegraphics[width=\linewidth]{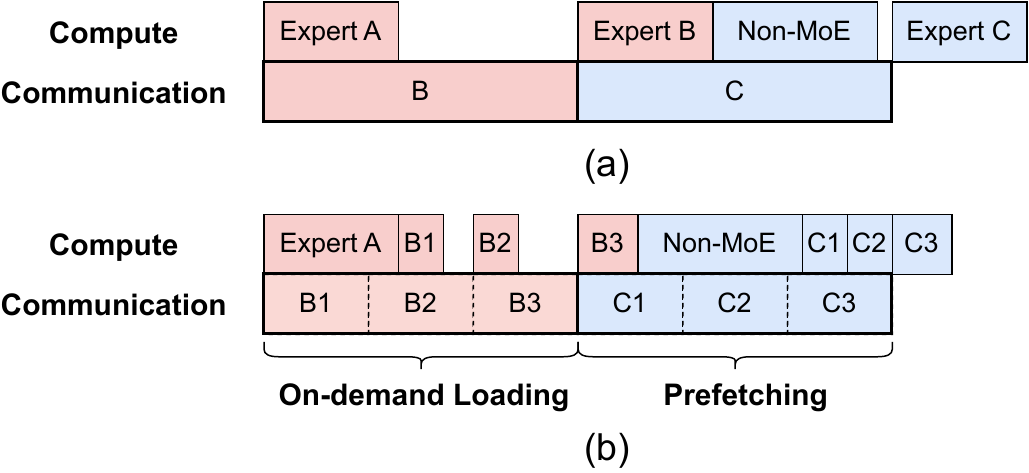}
    \caption{(a) Timeline of expert-wise scheduling; (b) timeline of tile-wise scheduling. Tile-wise scheduling enables to further overlap the computation and communication for better efficiency.}
    \label{fig:tile}
\end{figure}

To address the delay associated with the on-demand loading of experts, which cannot be processed until fully loaded into the GPU, our system employs a strategic tiling approach. As illustrated in Figure~\ref{fig:tile}, this strategy involves dividing each expert into smaller tiles. Each tile is loaded and processed as soon as it becomes available, enabling computations to significantly overlap with data loading. This method effectively minimizes the delays typically caused by data transfers, thereby accelerating the overall computation process and enhancing system throughput.


\begin{algorithm}[!tb]
\caption{AdapMoE System Workflow}
\label{algorithm}
\begin{algorithmic}[1]
\State \textbf{Data Structures:}
\State $E \gets \text{Set of required experts for the current layer}$
\State $r \gets \text{Experts to prefetch for subsequent layers}$
\State $h \gets \text{Activations}$
\State \textbf{end Data Structures}
\State
\Function{ComputeStream}{}
        \State $E, r \gets \Call{AdaptiveGates}{h}$
        \State $to\_transfer \gets E.in\_CPU \cup r$ 
        \State \Call{Enqueue}{transfer\_queue, to\_transfer}
        \State \Call{Notify}{comm\_stream}
        \State \Call{Enqueue}{compute\_queue, $E.in\_GPU$}
        \State \Call{Enqueue}{compute\_queue, $E.in\_CPU$}
        \State \Call{ProcessQueue}{compute\_queue, activations}
\EndFunction
\Function{CommStream}{}
    \While{\texttt{true}}
        \If{\Call{NotEmpty}{transfer\_queue}}
            \State $expert \gets \Call{Dequeue}{transfer\_queue}$
            \ForAll{$tile$ in $expert$}
                \State \Call{Transfer}{tile}
                \State \Call{NotifyTile}{tile}
            \EndFor
        \EndIf
    \EndWhile
\EndFunction
\Function{ProcessQueue}{queue,h}
    \ForAll{$expert$ in $queue$}
        \If{$expert$ in GPU}
            \State \Call{Compute}{expert,h}
        \Else
            \State \Call{Split}{expert}
            \ForAll{$tile$ in $expert$}
                \State \Call{WaitForTile}{tile}
                \State \Call{Compute}{tile,h}
            \EndFor
            \State $expert.in\_GPU$ = True
        \EndIf
    \EndFor
\EndFunction
\end{algorithmic}
\end{algorithm}
\section{Experimental Results}
\label{sec:exp}

\subsection{Experiment setup}

\begin{figure}[!tb]
    \centering
    \includegraphics[width=\linewidth]{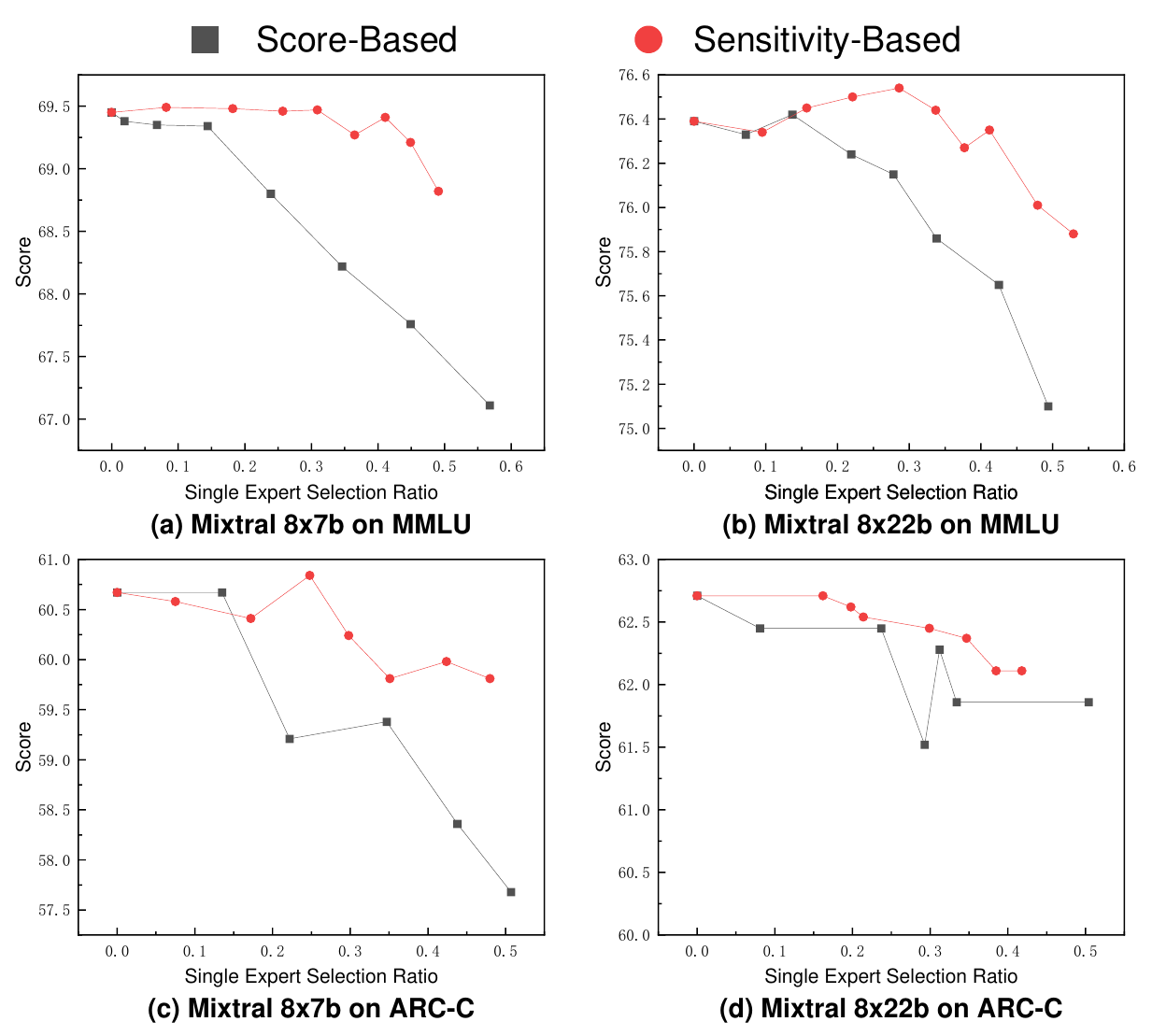}
    \caption{Accuracy comparison between the adaptive sensitivity-based gating and the score-based gating for different MoE models and NLP benchmarks.}
    \label{fig:acc}
\end{figure}

\textbf{Models.} We utilize Mixtral \cite{jiang2024mixtral} as the baseline MoE for our evaluations, a widely-used open-source MoE model. We test the 8x7b and 8x22b models under different quantization configurations to demonstrate the scalability of AdapMoE under different model sizes.

\textbf{Datasets.} We use two famous LLM benchmark datasets to evaluate the accuracy of adaptive gating, including MMLU \cite{hendrycks2020mmlu} and ARC-Challenge \cite{clark2018arc}.  Additionally, for testing generation speed, we sample prompts from the MT Bench \cite{zheng2024mtbench} to create realistic usage scenarios.

\textbf{Platforms.} We test the Mixtral-8x7b model on RTX 4090 and A6000, and Mixtral-8x22b model on A6000. We test different sizes of total cached experts to demonstrate the scalability of AdapMoE.

\subsection{Accuracy Comparison}
In this section, we quantify the impact of our adaptive gating mechanism on MoE's model accuracy. We sample different threshold to get different single expert activation ratios. All the models are quantized into 4bit in the experiments. We test the model accuracy on two famous LLM benchmark datasets, MMLU and ARC-Challenge.

Figure~\ref{fig:acc} illustrates the model accuracy differences when employing score-based  \cite{li2023adaptive} versus sensitivity-based adaptive gating mechanisms under varying single expert activation ratios. Notably, sensitivity-based adaptive gating demonstrates a higher tolerance for increased single expert activation ratios without a corresponding decrease in accuracy. This robustness suggests that sensitivity-based gating more effectively leverages the most relevant experts, maintaining high performance even as the selection becomes more concentrated.

\subsection{Performance Comparison}

\begin{figure}[!tb]
    \centering
    \includegraphics[width=\linewidth]{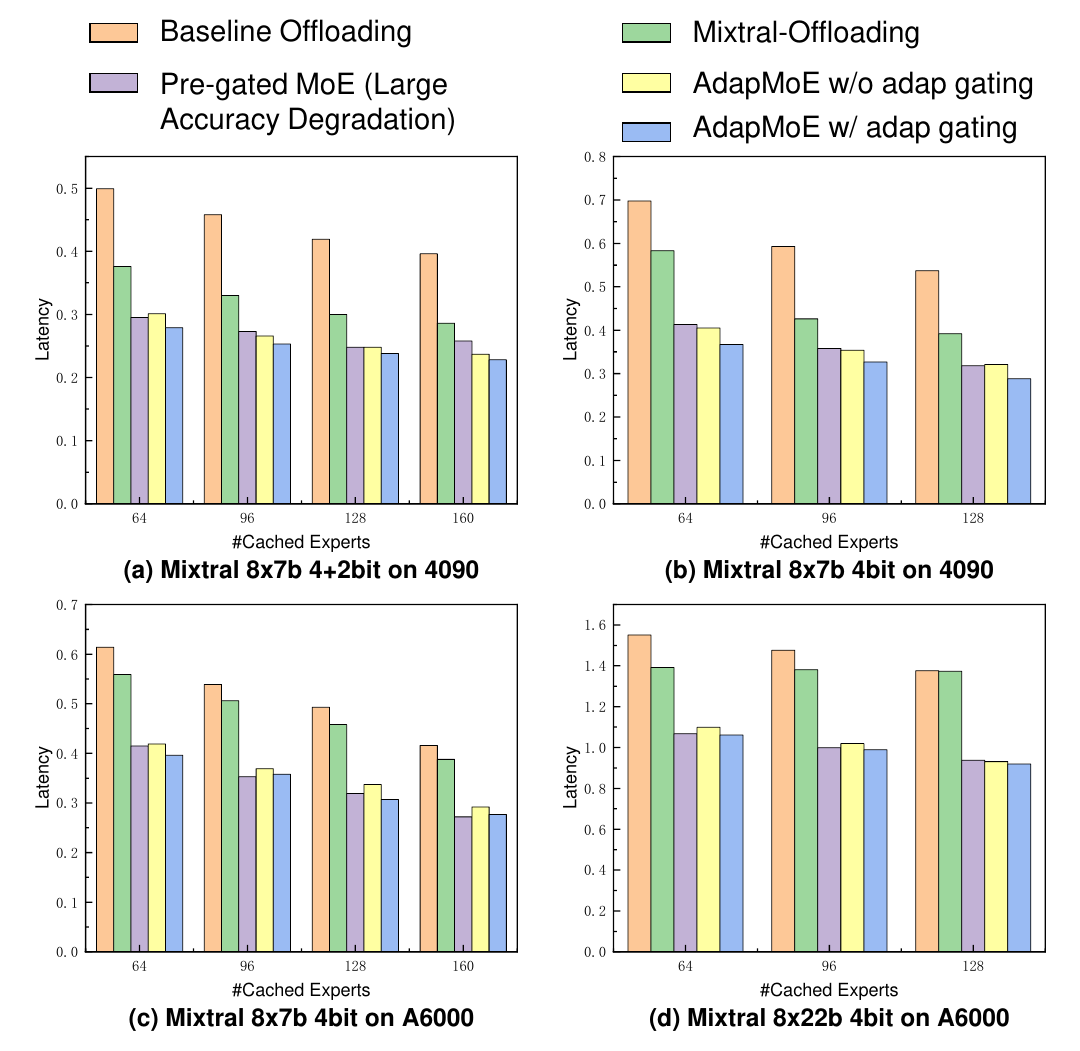}
    \caption{Inference speed comparison between AdapMoE and the baselines for different quantization configurations, model sizes, cached experts sizes, and platforms.}
    \label{fig:performance}
\end{figure}

We evaluate our framework against baseline offloading algorithms, Mixtral-offloading\footnote{Mixtral-offloading will not store all of the experts in CPU. We modify this configuration for fair comparison. The results of Mixtral-offloading in our experiments is 2x faster than its open-source version.}
and Pre-gated MoE.
Baseline offloading algorithm will load/offload the whole layer as adopted in Deepspeed \cite{rajbhandari2021zero}, Flexgen \cite{sheng2023flexgen}, etc.
Pre-gated MoE is the SOTA MoE inference mechanism.
We implement its algorithm by modifying the structure of Mixtral to prefetch and select experts based on the activations from the previous layer. All the methods take LRU as its cache elimination algorithm.

In our evaluations of AdapMoE, we conducted tests with and without the adaptive gating mechanism. The version without adaptive gating was configured to ensure identical output consistency with the baseline offloading and Mixtral-offloading. For the version equipped with adaptive gating, we choose a conservative single expert activation ratio of 24\%. This setting was chosen to safeguard the model's accuracy for fair comparison.

The mixtral models are quantized into 4+2 bit and 4bit using hqq framework \cite{badri2023hqq}. 4+2 bit will quantize the attention blocks into 4bit and MoE blocks into 2bit for efficient inference adopted by mixtral-offloading.

As shown in figure~\ref{fig:performance}, AdapMoE significantly reduces latency by an average 1.35$\times$ vs. Mixtral-Offloading. 
AdapMoE also exhibits comparable latency to Pre-gated MoE. The version of AdapMoE without adaptive gating matches the performance of Pre-gated MoE, while the version with gating consistently surpasses it, achieving better results without the need for additional finetuning or training.
Compared with Pre-gated MoE, this performance improvement is largely due to the implementation of adaptive prefetching and caching strategies. These strategies will prefetch experts for the next two or three layers and allocate additional cache size to more crucial layers, optimizing resource use and minimizing latency.

\subsection{Ablation Study}

We further explore how the components of our method contribute to the result. We take Mixtral-8x7b 4bit on 4090 with 128 cached experts as an example and show the impact of the proposed speedup techniques on the per-token latency. The baseline technique is our modified version of Mixtral-offloading on our framework, which is 2$\times$ faster than its open-source version.

As shown in table~\ref{tab:ablation}, each component of AdapMoE contributes to reducing the latency. The combined application of all techniques can lead to 1.36$\times$ speedup.

\begin{table}
\centering
\caption{MoE inference speedup breakdown of proposed techniques.}
\label{tab:ablation}
\begin{tabular}{c|cc} 
\hline \hline
Technique & latency(s) & speedup  \\ 
\hline
baseline                  & 0.392   &          \\
basline+gating            & 0.313   & 1.25$\times$     \\
baseline+prefetch         & 0.322   & 1.22$\times$     \\
baseline+gating+cache     & 0.305   & 1.29$\times$     \\
baseline+prefetch+cache   & 0.321   & 1.22$\times$     \\
baseline+gating+prefetch  & 0.308   & 1.27$\times$     \\
all                       & 0.288   & \textbf{1.36}$\times$     \\
\hline \hline
\end{tabular}
\end{table}

\subsection{Discussions}

\begin{figure}[!tb]
    \centering
    \includegraphics[width=\linewidth]{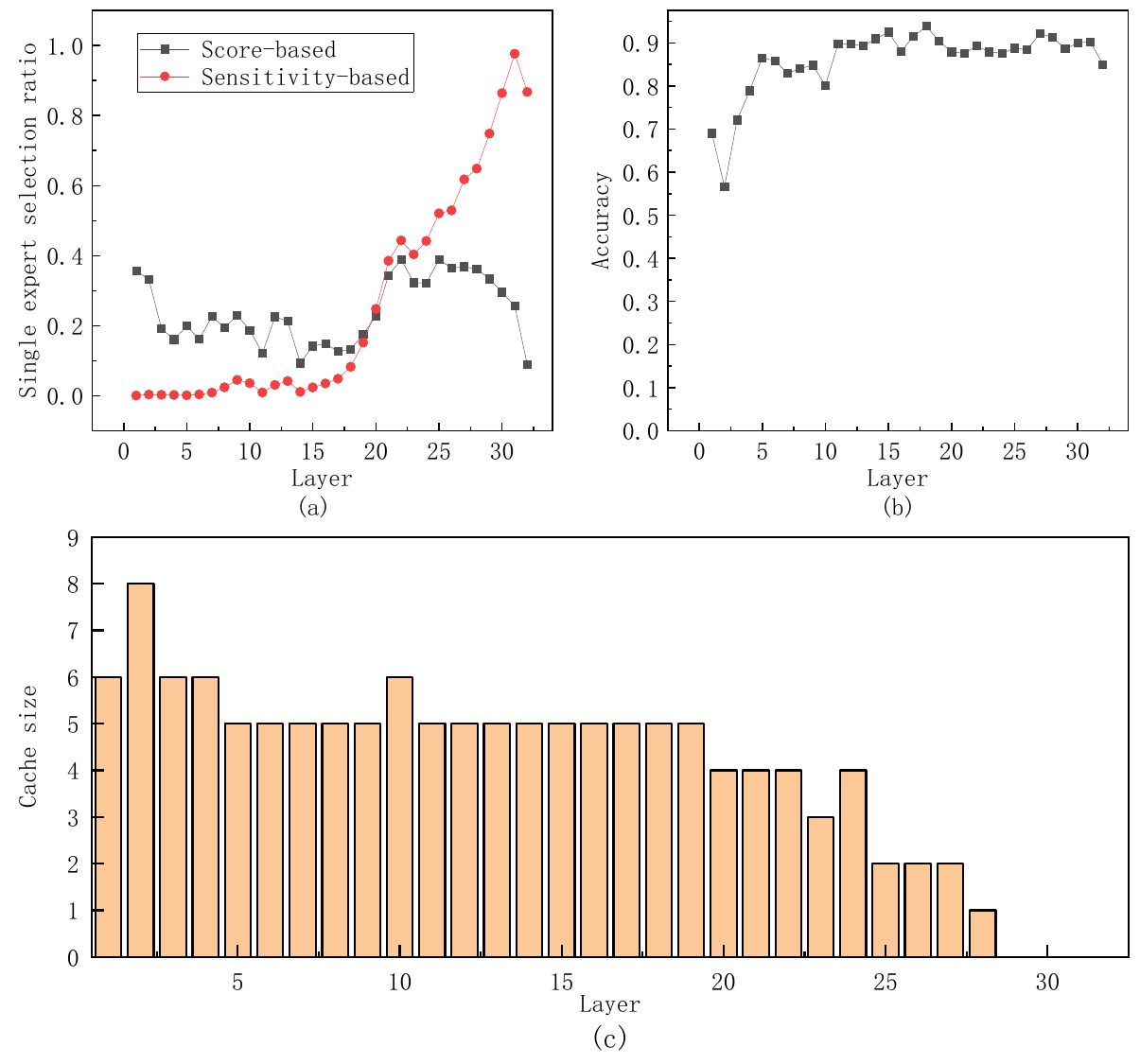}
    \caption{(a) Single expert selection ratios; (b) expert prefetch prediction accuracy of different layers; (c) expert cache allocation for different layers.}
    \label{fig:discussion}
    \vspace{-2mm}
\end{figure}

\textbf{Sensitivity.} Figure~\ref{fig:discussion}(a) presents the single expert activation ratios of score-based gating and sensitivity-based gating. Earlier layers are more sensitive to perturbation. In comparison to score-based adaptive gating, AdapMoE tends to activate more experts in early layers, which contributes to enhanced overall accuracy.

\textbf{Prefetch accuracy.} Figure~\ref{fig:discussion}(b) illustrates the prefetching accuracy across different layers. The prediction for the first layer is derived from the trained predictive gates, while for subsequent layers, it is based on the reuse of existing gates. The graph demonstrates that most layers maintain a high level of expert activation prediction accuracy, achieving rates around 90\%. 

\textbf{Cache allocation.} Figure~\ref{fig:discussion}(c) shows the cache size allocation for Mixtral 8x7b with total cache size 128. More experts of the early layers will be stored in GPU memory, for the early layers are more sensitivity and harder to be prefetched.
\section{Conclusion}
This paper introduces AdapMoE, an algorithm-system co-design for scalable and efficient MoE inference on edge devices. By implementing adaptive expert gating, prefetching, and caching within a unified framework, AdapMoE addresses the challenges of high latency and substantial on-demand loading overheads commonly associated with MoE models. Our evaluations across various hardware platforms demonstrate that AdapMoE not only reduces the activation of experts by 25\% but also achieves a 1.35x speedup in inference times, all without compromising the accuracy of the model.
\newpage

\bibliographystyle{ACM-Reference-Format}
\bibliography{reference/reference}

\end{document}